%% file: main.tex
\documentclass[10pt,twocolumn,letterpaper]{article}

\usepackage{cvpr}              %

\usepackage[pagebackref,breaklinks,colorlinks]{hyperref}

\usepackage{graphicx}
\usepackage{amsmath}
\usepackage{amssymb}
\usepackage{booktabs}
\usepackage{multirow, makecell}
\usepackage{graphicx}
\usepackage{adjustbox}
\usepackage[accsupp]{axessibility}

\usepackage{algorithm}
\usepackage{algpseudocode}

\usepackage[capitalize]{cleveref}
\crefname{section}{Sec.}{Secs.}
\Crefname{section}{Section}{Sections}
\Crefname{table}{Table}{Tables}
\crefname{table}{Tab.}{Tabs.}

\def\confName{CVPR}
\def\confYear{2023}

\newcommand{\ourmodel}{{UltraLiDAR}}
\newcommand{\twostage}{{Two-stage PIXOR}}

\usepackage{color}
\definecolor{alizarin}{rgb}{0.82, 0.1, 0.26}

\makeatletter
\g@addto@macro\@maketitle{
\vspace{-1.0em}
\begin{figure}[H]
\setlength{\linewidth}{\textwidth}
\setlength{\hsize}{\textwidth}
\centering
\includegraphics[trim={0cm, 0cm, 0cm, 0cm},clip,width=\textwidth]{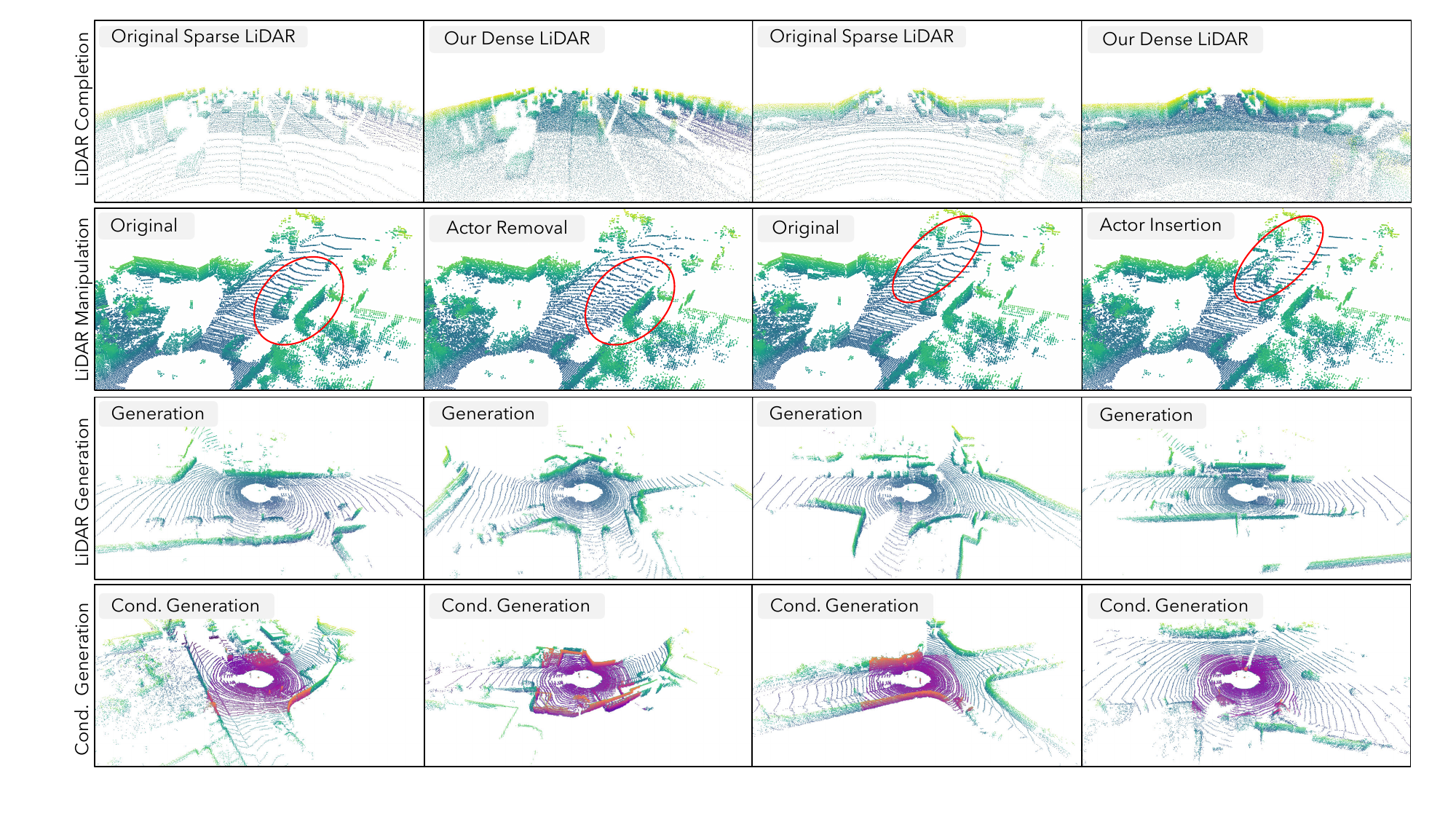}
\vspace{-6.5mm}
\caption{%
UltraLiDAR learns discrete representations from large-scale LiDAR point clouds and conducts realistic, scalable and controllable LiDAR completion and generation.
\textbf{Top row: } Sparse-to-dense LiDAR completion;
\textbf{Second row: }  Controllable manipulation of real LiDAR with actor removal and insertion;
\textbf{Third row: } Diverse LiDAR generation with realistic global structure and fine-grained details;
\textbf{Bottom row: } Conditional scene generation with \textcolor{alizarin}{partially observed point clouds} (highlighted in red). Please see supp. for more examples. }\label{fig:teaser}
\end{figure}
}
\makeatother

\begin{document}

\title{Learning Compact Representations for LiDAR Completion and Generation}

\author{Yuwen Xiong$^{1, 2}$ \quad\quad Wei-Chiu Ma$^{1, 3}$ \quad\quad  Jingkang Wang$^{1, 2}$ \quad\quad  Raquel Urtasun$^{1, 2}$\\
$^{1}$Waabi \quad $^{2}$University of Toronto \quad $^{3}$Massachusetts Institute of Technology\\
{\tt\small \{yxiong,wma,jwang,urtasun\}@waabi.ai}
}
\maketitle

\input{abstract}

\input{intro-new}

\input{related}

\input{method-new}

\input{experiments-new}

\input{conclusion}

\input{acknowledgement}

\bibliographystyle{ieee_fullname}
\bibliography{egbib}
\clearpage
\addcontentsline{toc}{section}{Appendix}
\section*{Appendix}
\appendix
\renewcommand{\thetable}{A\arabic{table}}
\renewcommand{\thefigure}{A\arabic{figure}}
\input{supp.tex}

\end{document}

%% file: abstract.tex
\begin{abstract}
LiDAR provides accurate geometric measurements of the 3D world.
Unfortunately, dense LiDARs are very expensive and the point clouds captured by low-beam LiDAR are often sparse.
To address these issues, we present UltraLiDAR, a data-driven framework for scene-level LiDAR completion, LiDAR generation, and LiDAR manipulation.
The crux of UltraLiDAR is a compact, discrete representation that encodes the point cloud's geometric structure, is robust to noise, and is easy to manipulate.
We show that by aligning the representation of a sparse point cloud to that of a dense point cloud, we can densify the sparse point clouds as if they were captured by a real high-density LiDAR, drastically reducing the cost.
Furthermore, by learning a prior over the discrete codebook, we can generate diverse, realistic LiDAR point clouds for self-driving.
We evaluate the effectiveness of UltraLiDAR on sparse-to-dense LiDAR completion and LiDAR generation.
Experiments show that densifying real-world point clouds with our approach can significantly improve the performance of downstream perception systems.
Compared to prior art on LiDAR generation, our approach generates much more realistic point clouds. According to A/B test, over 98.5\% of the time human participants prefer our results over those of previous methods.
The project page is available at \url{https://waabi.ai/ultralidar/}.
\end{abstract}

%% file: intro-new.tex
\vspace{-4mm}
\section{Introduction}
\label{sec:intro}

Building a robust 3D perception system is key to bringing self-driving vehicles into our daily lives. %
To effectively perceive their surroundings, existing autonomous systems primarily exploit LiDAR as the major sensing modality, since it can capture well the 3D geometry of the world.
However, while LiDAR provides accurate geometric measurements, it comes with two major limitations: (i) the captured point clouds are inherently sparse; and (ii) the data collection process is difficult to scale up.

\vspace{-4mm}
\paragraph{Sparsity:}
Most popular self-driving LiDARs are time-of-flight and scan the environment by rotating emitter-detector pairs (\ie, beams) around the azimuth.
At every time step, each emitter emits a light pulse which travels until it hits a target, gets reflected, and is received by the detector.
Distance is measured by calculating the time of travel.
Due to the design, the captured point cloud density inherently decreases as the distance to the sensor increases.
For distant objects, it is often that only a few LiDAR points are captured, which greatly increases the difficulty for 3D perception.
The sparsity problem becomes even more severe under poor weather conditions~\cite{xu2021spg}, or when LiDAR sensors have fewer beams~\cite{caesar2020nuscenes}.
One ``simple'' strategy is to increase the number of LiDAR beams.
However, 128-beam LiDAR sensors are much more expensive than their 64/32-beam counterparts, not to mention that 512-beam LiDAR does not exist yet.

\vspace{-4mm}
\paragraph{Scalability:}
Training and testing perception systems in diverse situations are crucial for developing robust autonomous systems.
However, due to their intricate design, LiDARs are much more expensive than cameras.
A commercial 64-beam LiDAR usually costs over 25K USD.
The price barrier makes LiDAR less accessible to the general public and restricts data collection to a small fraction of vehicles that populate our roads, significantly hindering scaling up.
One way to circumvent the issue is to leverage existing LiDAR simulation suites to generate more data.
While the simulated point clouds are realistic, these systems typically require one to manually create the scene or rely on multiple scans of the real world in advance, making such a solution less desirable.

With these challenges in mind, we propose \ourmodel{}, a novel framework for LiDAR completion and generation.
The key idea is to learn a \emph{compact, discrete} 3D representation (codebook) of LiDAR point clouds that encodes the geometric structure of the scene and the physical rules of our world (\eg, occlusion).
Then, by aligning the representation of a sparse point cloud to that of a dense point cloud,
we can densify the sparse point cloud as if it were captured by a high-density LiDAR (\eg, 512-beam LiDAR).
Furthermore, we can learn a prior over the discrete codebook and generate novel, realistic driving scenes by sampling from it; we can also manipulate the discrete code of the scene and produce counterfactual scenarios, both of which can drastically improve the diversity and amount of LiDAR data. Fig. \ref{fig:teaser} shows some example outputs of UltraLiDAR.

We demonstrate the effectiveness of UltraLiDAR on two tasks: sparse-to-dense LiDAR completion and LiDAR generation.
For LiDAR completion, since there is no ground truth for high-density LiDAR,
we exploit the performance of downstream perception tasks as a ``proxy'' measurement.
Specifically, we evaluate 3D detection models on both sparse and densified point clouds and measure the performance difference. Experiments on Pandaset \cite{xiao2021pandaset} and KITTI-360 \cite{Liao2021ARXIV} show that our completion model can generalize across datasets and the densified results can significantly improve the performance of 3D detection models.
As for LiDAR generation, we compare our results with state-of-the-art LiDAR generative models \cite{caccia2019deep,zyrianov2022lidargen}. Our generated point clouds better match the statistics of those of ground truth data. We also conducted a human study where participants prefer our method over prior art over 98.5\% (best 100\%) of the time; comparing to ground truth, our results were selected 32\% of the time (best 50\%).

To summarize, the main contributions of this paper are:
\begin{enumerate}
\vspace{-2mm}
  \item We present a LiDAR representation that can effectively capture data priors and enable various downstream applications, \eg, LiDAR completion and generation.
  \vspace{-2mm}
  \item We propose a sparse-to-dense LiDAR completion pipeline that can accurately densify sparse point clouds and improve the performance and generalization ability of trained detection models across datasets.
  \vspace{-2mm}
  \item We develop an (un)conditional LiDAR generative model that can sythesize high-quality LiDAR point clouds and supports various manipulations.
\end{enumerate}

%% file: related.tex
\vspace{-3mm}
\section{Related Work}

\input{figures/model.tex}

\paragraph{LiDAR / Depth Completion:} Sparse LiDAR/depth completion is a beneficial task for many robotic applications as it can provide more accurate geometry for holistic scenes compared to images.
LiDAR scene completion has been studied and attracted the computer vision community's attention in recent years~\cite{behley2019semantickitti,song2017semantic,cheng2020s3cnet,roldao2020lmscnet,yan2021sparse,rist2021semantic}, while the progress has been pushing forward, most of them focus on directly predicting semantic labels for the completed coarse voxels without fine-grained geometry information, thus no other downstream tasks are benefited.
More recently, SPG~\cite{xu2021spg} was proposed for unsupervised domain adaptation for 3D object detection via foreground point completion. However, its completion target is generated by heuristics, thus an improved holistic understanding of the scene is hard to achieve.
Depth completion~\cite{Uhrig2017THREEDV} instead solves the problem by relying on a sparse lidar sweep and a paired RGB image to compute a 2.5D pseudo-depth image.
However, this problem is ill-posed as monocular images contain ambiguities and the field of view is also limited.
Thus, the accuracy is hard to improve, and it's difficult to reconstruct whole scenes at scale.
Our model instead tackles this problem from a representation learning perspective.
It captures the data prior in advance with robust discrete representations thus generalizing better.
It also completes the full scene with a higher-beam LiDAR pattern, which is generic to various downstream tasks.

\vspace{-3mm}
\paragraph{3D Generation:} Inspired by the tremendous progress in 2D image generation~\cite{goodfellow2020generative,karras2019style,dhariwal2021diffusion}, 3D content generation has attracted more and more attention in recent years.
Existing works demonstrated the high-quality generation in different representations including point cloud~\cite{caccia2019deep,yang2019pointflow,achlioptas2018learning,zhou20213d}, voxel grid~\cite{wu2016learning,lunz2020inverse,smith2017improved,girdhar2016learning,henzler2019escaping}, mesh~\cite{gao2022get3d,groueix2018papier,gkioxari2019mesh,shen2021deep} or implicit geometry~\cite{eg3d,graf,pigan}.
However, these works usually focus on the object level and cannot handle large scenes (due to non-compact representation, large memory footprint and limited model capacity).
To enable larger-scale scene synthesis, recent works either define a more compact output space or procedurally generate the scenes~\cite{peng2014computing,nauata2020house}.
Unfortunately, the generated scenes are still quite limited (\eg, indoor environments) and there are only few works focusing on large-scale scene generation for self-driving.
In this paper, we focus on the generation of LiDAR point clouds that are highly unstructured, sparse, and non-uniform.
~\cite{caccia2019deep} and~\cite{sallab2019lidar} use the variational autoencoder (VAE) or generative adversarial network (GAN) on LiDAR point clouds but the generation realism is quite limited.
Most recently, researchers propose a novel score-matching diffusion model for higher-quality LiDAR generation~\cite{zyrianov2022lidargen} but keep only limited local geometry details and global scenario structures.
In contrast, we present a generic framework to learn discrete representations to better maintain the structure and semantic information for a more realistic and controllable generation.

\vspace{-3mm}
\paragraph*{Learning Discrete Representations:}
VQ-VAE~\cite{van2017neural} proposed to learn discrete representations by compressing images into discrete latent space.
It primarily consists of three components: an encoder that learns to map the input image into latent feature maps; a codebook that contains a set of learnable latent embeddings where the feature maps are quantized via nearest neighbor lookup; and a decoder that reconstructs the input image from the quantized code.
~\cite{razavi2019generating} used a multi-scale hierarchical architecture to capture local and global information in separate codebooks.
Based on the learned codebook and decoder in VQ-VAE, VQ-GAN~\cite{esser2021taming} proposed to use of a transformer model for image generation.
More recently, MaskGIT~\cite{chang2022maskgit} shows that the learned codebook and the decoder are capable of generating impressive realistic RGB images by using mask modeling with bi-directional transformers.
However, all existing works focus on 2D natural images and it is non-trivial to handle sparse, unstructured LiDAR points.
To our best knowledge, this paper is the first work that conducts discrete representation in the 3D domain and achieves state-of-the-art performance in large-scale LiDAR scene completion and generation.

%% file: figures/model.tex
\begin{figure*}
    \centering
    \vspace{-5mm}
    \includegraphics[trim={0cm, 0.35cm, 0cm, 0.0cm},clip,width=\textwidth]{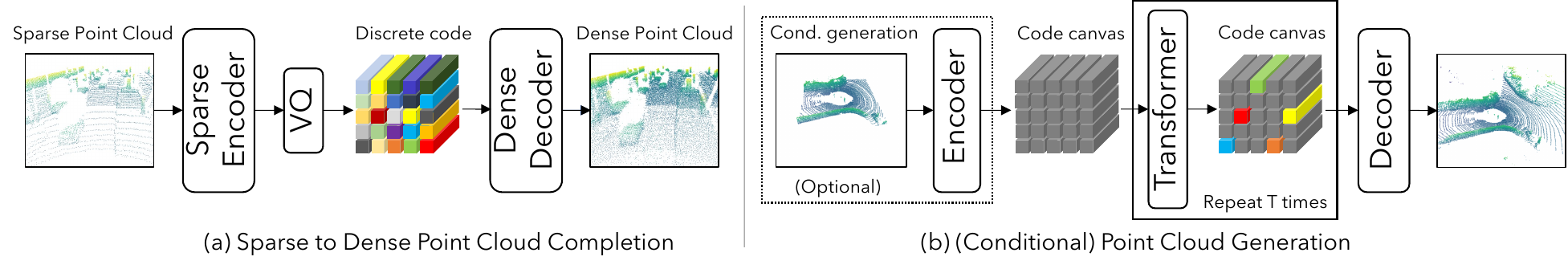}
    \vspace{-6mm}
    \caption{\textbf{Overview of \ourmodel{} pipeline.}
    For (a) LiDAR completion, the sparse encoder maps the sparse point cloud to discrete codes, and the dense decoder reconstructs dense data from them;
    For (b) LiDAR generation, the transformer model starts from a blank canvas or canvas with codes mapped from the partial observations; and iteratively predicts and updates the missing parts.
    The decoder produces the LiDAR output given the predicted code as the generation results.
    }\label{fig:model}
    \vspace{-5mm}
\end{figure*}

%% file: method-new.tex
\vspace{-1mm}
\section{UltraLiDAR}
\label{sec:method}
\vspace{-1mm}

\input{figures/sparse_to_dense.tex}

In this paper, we seek to learn a compact 3D representation of scene-level LiDAR point clouds to enable a series of downstream applications, such as sparse-to-dense LiDAR completion, (un)conditional LiDAR generation, and LiDAR manipulation.
Based on the observation that vector quantized (VQ) representations~\cite{van2017neural,razavi2019generating,esser2021taming} are robust to noise, easy to manipulate, and naturally compatible with generative models, we propose to learn a discrete codebook for LiDAR point clouds and build our model on top of it.

We start by reviewing the basics of vector-quantized variational autoencoder (VQ-VAE)~\cite{van2017neural}, a building block of our approach.
Then we showcase how to exploit similar concepts to encode 3D point clouds into a discrete codebook.
Finally, we discuss how to exploit the discrete representation for different tasks.

\vspace{-1mm}
\subsection{Discrete Representations for LiDAR}
\label{sec:vqvae}
\paragraph{VQ-VAE revisit:}
The goal of VQ-VAE is to learn a discrete latent representation that is expressive, robust to noise, and compatible with generative models.
VQ-VAE consists of three parts:
(i) an encoder $E$ that encodes the input signal, which for simplicity we assume to be an image,
$\mathbf{x} \in \mathbb{R}^{H\times W\times 3}$ to a continuous embedding map $\mathbf{z} = E(\mathbf{x}) \in \mathbb{R}^{h\times w\times D}$,
(ii) an element-wise quantization function $q$ that maps each embedding to its closest learnable latent code $\mathbf{e}_k \in \mathbb{R}^D$, with $k=1, ..., K$,
and (iii) a decoder $G$ that takes as input the quantized representation $\hat{\mathbf{z}} = q(\mathbf{z})$ and outputs the reconstructed image $\hat{\mathbf{x}} = G(\hat{\mathbf{z}})$.
The whole model can be trained end-to-end by minimizing:
\begin{align}
\mathcal{L}_{\text{vq}} = \lVert\mathbf{x} - \hat{\mathbf{x}}\rVert^2_2 + \lVert\text{sg}[E(\mathbf{x})] - \hat{\mathbf{z}}\rVert^2_2 + \lVert\text{sg}[\hat{\mathbf{z}}] - E(\mathbf{x})\rVert^2_2,
\label{eq:vqvae}
\end{align}
where {sg[$\cdot$]} denotes the stop-gradient operation.

The limited number of discrete codes $\mathbf{e}$ stabilizes %
the input distribution of the decoder during training; it also forces the codes to capture meaningful, re-usable information as the decoder can no longer ``seek shortcut'' from the continuous signals for the reconstruction task.
VQ-VAE has enjoyed great success across different modalities, such as natural images \cite{esser2021taming,chang2022maskgit}, audio \cite{van2017neural} and 2.5D images \cite{shen2022sgam}.
In this work, we further extend it to learn discrete representations for 3D LiDAR point clouds.

\vspace{-3mm}
\paragraph{VQ-VAE for LiDAR:}
We aim to learn a discrete codebook that can effectively represent a set of LiDAR point clouds.
Directly applying VQ-VAE, however, is challenging,
since the fixed set of discrete latents will have to model
point clouds that live in a continuous 3D space, and deal with the fact that each point cloud may have a different number of points.
To address these issues, we propose to voxelize the point clouds and instead infer  whether each voxel is occupied or not.
By grounding the point clouds with a pre-defined grid (similar to the 2D pixel grid of images), we can foster the discrete codebook to learn the overall structure rather than the minor 3D positional variations.
This representation also can  naturally handle the varying number of points.
While we may sacrifice some precision during the voxelization process, the impact is negligible for both LiDAR completion and generation (see Sec. \ref{sec:exp}).

Now that we have a voxelized point cloud, the next step is to design the encoder $E$ and the decoder $G$.
For large scenes with high resolution, 3D convolution becomes very expensive since we need to infer the occupancy of each voxel densely.
We thus convert the input to  Birds-Eye-View (BEV) images by treating the height dimension of the voxel grids as feature channel $C$ and then adopt 2D convolutions instead.
In this case, we can process 3D LiDAR data just like 2D images; we can also exploit existing model architectures designed for 2D images directly.
We note that such BEV images have been widely adopted in the context of self-driving perception~\cite{chen2017multi,zhang2018efficient},
since they encode rich geometric information.
The output of the decoder is a logit grid $\hat{\mathbf{x}} \in \mathbb{R}^{H\times W\times C}$. It can be further converted to a binary voxel grid $\hat{\mathbf{x}}^\text{bin} \in \{0, 1\}^{H\times W\times C}$ through gumbel softmax \cite{jang2016categorical}.

Finally, we train our LiDAR VQ-VAE model with Eq.~\ref{eq:vqvae}, except that we replace the $\ell_2$ reconstruction loss with a binary (occupied or not) cross-entropy loss.
To improve the realism of the reconstruction, we further adopt a pre-trained voxel-based detector $V$ and measure the feature difference, similar to perpectual loss \cite{johnson2016perceptual}. Our full loss is:
\begin{align}
\mathcal{L}_{\text{feat}} = \mathcal{L}_{\text{vq}} + \lVert V_{\mathrm{b}}(\mathbf{x}) - V_{\mathrm{b}}(\hat{\mathbf{x}}^\text{bin}) \rVert_2^2.
\end{align}
$V_{\mathrm{b}}$ denotes the feature from the last backbone layer of $V$.

\vspace{-3mm}
\paragraph{LiDAR manipulation:}
Once we train the model, we can easily manipulate arbitrary LiDAR point clouds by editing their corresponding latent codes.
Since objects are spatially apart in 3D, the model can dedicate specific codes for them.
We can thus identify the codes for objects of interest (\eg, vehicles) and insert/remove them into/from the scene.
As shown in Sec. \ref{sec:exp}, we can populate many vehicles on the street and create counterfactual scenarios. %

\input{figures/kitti360_comparison.tex}

\subsection{LiDAR Completion}
\label{sec:method-completion}
Given a dataset of paired, voxelized LiDAR point clouds
$\{(\mathbf{x}^{\text{sp}}_{1}, \mathbf{x}^{\text{den}}_{1}), ..., (\mathbf{x}^{\text{sp}}_{N}, \mathbf{x}^{\text{den}}_{N})\}$, the goal of LiDAR completion is to learn a function $f$ that maps a sparse LiDAR point cloud $\mathbf{x}^{\text{sp}}$ to its dense counterpart $\mathbf{x}^{\text{den}}$.

One straightforward strategy is to leverage the pairwise supervision and directly learn a pix2pix-style network \cite{isola2017image} but on voxels.
While this model performs well within the same data distribution, it degrades significantly when the input distribution shifts (\eg, when the sparse LiDAR point clouds come from different datasets).
We conjecture that this is because there is no restriction on the learned representation, and the model learns to ``cheat'' by relying on non-generalizable details rather than the overall structure.

In this section, we investigate the use of discrete representations to alleviate the issue. We first learn a discrete codebook $\{\mathbf{e}^{\text{den}}_1, ..., \mathbf{e}^{\text{den}}_K\}$, an encoder $E^{\text{den}}$, and a decoder $G^{\text{den}}$ for each dense LiDAR point cloud $\mathbf{x}^{\text{den}}$ using the approach in Sec. \ref{sec:vqvae}.
Then we learn a separate encoder $E^{\text{sp}}$ that maps each sparse LiDAR point cloud $\mathbf{x}^{\text{sp}}$ to the same feature space $\mathbf{z}^{\text{sp}} = E^{\text{sp}}(\mathbf{x}^{\text{sp}})$ and quantize it with the dense discrete representation $\mathbf{e}^{\text{den}}$. Finally, we decode the quantized representation $\hat{\mathbf{z}}^{\text{sp}} = q(\mathbf{z}^{\text{sp}})$ with the dense decoder $G^{\text{den}}$ and obtain a densified point cloud $\hat{\mathbf{x}}^{\text{sp-den}} = G^{\text{den}}(\hat{\mathbf{z}}^{\text{sp}})$.
With the help of the learned codebook, we can ensure the input to the dense decoder $G^{\text{den}}$ is in-domain and the reconstructed point clouds are  high-quality; we can also denoise the embedding from the encoder and better handle the variations and distribution shift within the input data. %

\vspace{-3mm}
\paragraph{Training:}
In practice, we jointly train the sparse encoder $E_{\text{sp}}$ and the dense VQ-VAE model, since it achieves slightly better performance than performing two-stage training.
We hypothesize this is because joint training allows the model to learn a codebook that is easy to decode, and achieves low quantization error for both encoders $E_{\text{sp}}$ and $E_{\text{den}}$.
We use the same loss functions in Sec. \ref{sec:vqvae} to train the whole model, except that the reconstructed target is always the dense point cloud (which we can obtain from the paired data).

\vspace{-1mm}
\subsection{LiDAR Generation}
\label{sec:method-generation}
As we have alluded to earlier, the learned discrete representations can be naturally combined with generative models. In this section, we describe in detail how we exploit the discrete codebook to generate high-fidelity LiDAR point clouds, both unconditionally and conditionally.

\vspace{-4mm}
\paragraph{Unconditional generation:}
Given the learned codebook $\mathbf{e}$ and the decoder $G$, the problem of LiDAR generation can be formulated as code map generation.
Instead of directly generating LiDAR point clouds, we first generate discrete code maps in the form of code indices.
Then we map the indices to discrete features by querying the codebook and decoding them back to LiDAR point clouds with the decoder.
Following Chang \etal~\cite{chang2022maskgit}, we adopt a bi-directional self-attention Transformer \cite{he2022masked,chang2022maskgit} to \emph{iteratively} predict the code map.
Specifically, we start from a blank canvas.
At each iteration, we select a subset of the predicted codes with top confidence scores and update the canvas accordingly.
With the help of the Transformer, we can aggregate context from the whole map and predict missing parts %
based on already predicted codes.
In the end, the canvas will be filled with predicted code indices, from which we can decode LiDAR point clouds.
We refer the reader to \cite{chang2022maskgit} for more details.

\vspace{-4mm}
\paragraph{Conditional generation:}
Our unconditional LiDAR generation pipeline can be easily extended to perform conditional generation.
Instead of starting the generation process from an empty canvas, one can simply start with a partially filled code map and let the Transformer predict the rest.
For instance, we can place \texttt{[CAR]} codes at regions of interest; and run the model multiple times.
We can then obtain different traffic scenarios with the pre-defined cars untouched.
Please refer to supp. material for how we identify the codes.
\vspace{-9mm}
\paragraph{Free space suppression sampling:}
Our iterative generation procedure can be viewed as a variant of coarse-to-fine generation. The codes generated during early iterations determine the overall structure, while the ones generated at the end are in charge of fine-grained details.
While this pipeline is effective for image generation \cite{chang2022maskgit}, it may lead to degenerated results when generating LiDAR point clouds.
One critical reason is that LiDAR point clouds are sparse, and a large portion of the scene is represented by the same \texttt{[BLANK]} codes. Since the \texttt{[BLANK]} codes occur frequently, the Transformer tends to predict them with high scores. If we naively sample the codes based on the output of the Transformer, we may fill most of the canvas with \texttt{[BLANK]} codes, and little structure will remain.
To address this issue, we suppress the \texttt{[BLANK]} codes during the early generation stages by setting their probability to 0. This ensures the model generates meaningful structures in the beginning.
We identify the \texttt{[BLANK]} codes by looking at the occurrence statistic of all codes across the whole dataset. We empirically select the top as \texttt{[BLANK]} codes.

\vspace{-3mm}
\paragraph{Iterative denoising:}
With free space suppression sampling, we can already obtain good results. However, the generated point clouds sometimes still contain high-frequency noise (\eg, there might be some floating points in the very far range).
To mitigate this issue, we randomly mask out different regions of the output LiDAR point clouds and re-generate them.
The intuition is that if we mask out a structured region, we can still recover it through the neighborhood context. However, if the masked region corresponds to pure noise that is irrelevant to the surrounding, it will likely be removed after multiple trials (since the model cannot infer it from the context).

\vspace{-3mm}
\paragraph{Training:}
We first encode all LiDAR point clouds into frozen discrete representations (code maps) learned in Sec. \ref{sec:vqvae}.
Then, at each training iteration, we randomly mask out a subset of codes.
Finally, we adopt the bi-directional Transformer to predict the correct code for those masked regions.
Since we have GT code map, we supervise the model with cross-entropy loss. See \cite{chang2022maskgit} for more details.

\vspace{-2mm}
\subsection{Implementation Details}

In this section, we discuss implementation details that are crucial for learning discrete LiDAR representations.

\vspace{-4mm}
\paragraph{Voxel sizes matter:} We set the input voxel size to be $15.625\times 15.625 \times 15 $ cm for x, y, z dimensions.
We find that the downsampling ratio when mapping the BEV image to the discrete code has a huge impact on both reconstruction and generation performance.
If the patch size that each discrete code represents is too large, a single fixed code does not have enough capacity to model the  variations (e.g., a small position/rotation shift of a car inside the patch).
We empirically find that downsampling $8\times$ achieves a good trade-off between preserving high-frequency information in the LiDAR data and maintaining high-level semantic meaning.
Thus the patch size is $8\times 8$, leading to $1.25 \times 1.25$ m on the spatial dimension that each code represents.

\input{tables/indomain_pandaset.tex}

\vspace{-4mm}
\paragraph{Model hyperparameters:} As a typical BEV image usually has a large spatial range, the resolution of the image and the number of codes are high.
We use Swin Transformer~\cite{liu2021swin} instead of the vanilla Vision Transformer~\cite{dosovitskiy2020image} to reduce the computational cost for our generative Transformer model.
It has 24 layers and 8 heads, and the embedding dimension is set to 512.
All other training hyper-parameters like optimizer settings and label smoothing are kept the same as in~\cite{chang2022maskgit}.
For simplicity, we use the same architecture in our VQ-VAE learning.
The encoder and decoder are both Swin Transformers with 12 layers, respectively.
We set the codebook size to 1024 with 1024 hidden dimensions for each code.

\vspace{-4mm}
\paragraph{Codebook (re)-initialization:}
We empirically find that the codebook can easily collapse (only a few codes are used) during training.
For better codebook learning, we use data-dependent codebook initialization.
Specifically, we use a memory bank to store the continuous embedding output from the encoders at each iteration;
and use K-Means centroids of the memory bank to initialize/reinitialize the codebook if the code utilization percentage is lower than a threshold (empirically, we define a code is not used for 256 iterations as ``dead code'' and set the threshold to be $50\%$).
During the first 2000 iterations of  training, we gradually shifted the decoder input from continuous to quantized embeddings as a warmup.
We find these strategies help  achieve good codebook learning and utilization.

%% file: figures/sparse_to_dense.tex
\begin{figure*}[t]
    \centering
    \vspace{-3mm}
    \includegraphics[trim={0cm, 0.0cm, 0cm, 0.0cm},clip,width=0.9\textwidth]{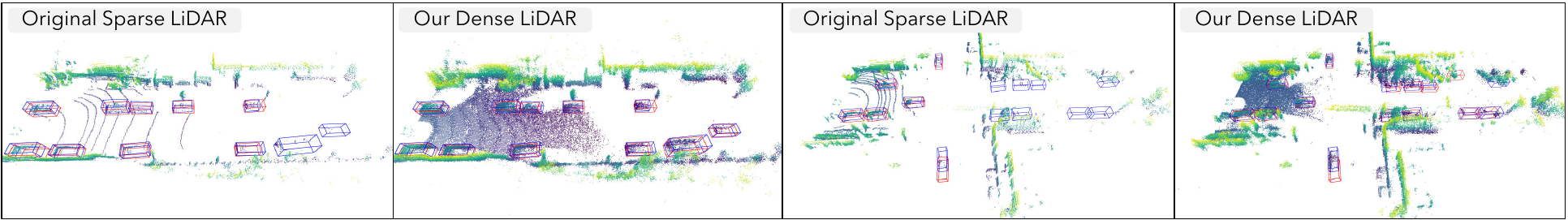}
    \vspace{-3mm}
    \caption{\textbf{LiDAR completion and 3D detection on Pandaset.} With densified point clouds, the detection model can identify more objects and reduce false negatives.
    We show \textcolor{red}{detection results} in red boxes and \textcolor{blue}{ground truth} in blue.
    The missing area far away from ego vehicle in the densified results is caused by uphill; we refer the reader to the supp. material for the explanation with camera visualization.
    }
    \label{fig:sparse_to_dense}
    \vspace{-3mm}
\end{figure*}

%% file: figures/kitti360_comparison.tex
\begin{figure*}[t]
    \centering
    \vspace{-7mm}
    \includegraphics[trim={0cm, 0.0cm, 0cm, 0.0cm},clip,width=0.9\textwidth]{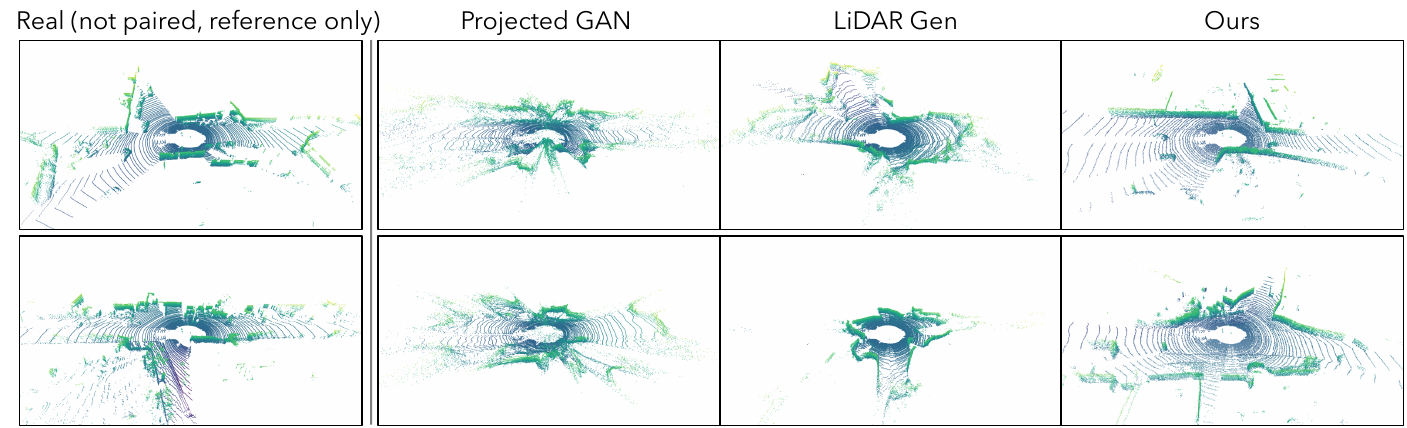}
    \vspace{-3mm}
    \caption{\textbf{Qualitative comparison against baselines on unconditional LiDAR generation.}
    We compare with two state-of-the-art LiDAR generation methods Projected GAN~\cite{sauer2021projected} and LiDARGen~\cite{zyrianov2022lidargen} and include real data for reference.
    Our model can generate results with more structured layouts and clearer beam patterns.}\label{fig:kitti360_comparison}
    \vspace{-5.5mm}
\end{figure*}

%% file: tables/indomain_pandaset.tex
\begin{table}[t]
    \centering
    \begin{adjustbox}{width=0.9\linewidth}
      \begin{tabular}{cc|cc|cc}
        \multirow{2}{*}{Model} & \multirowcell{2}{Sparse to\\ Dense} & \multicolumn{2}{c|}{\twostage{}} & \multicolumn{2}{c}{PointPillar} \\
      &  &$\mathrm{AP}_{\mathrm{BEV}}$ & $\mathrm{AP}_{\mathrm{3D}}$ & $\mathrm{AP}_{\mathrm{BEV}}$  & $\mathrm{AP}_{\mathrm{3D}}$ \\
        \hline
        Real / 64 & - & 79.3 & 62.2 & 75.5 & 62.3\\
        Sim / 512 & - & 78.1 & 57.7 & 70.0 & 55.5\\
        Sim / 512 & ContComp & 79.7 & 62.4 & 75.1 & 59.8\\
        Sim / 512 & Ours & {\bf 80.3} & {\bf 64.3} & {\bf 76.0} & {\bf 62.8}\\
        \hline
      \end{tabular}
    \end{adjustbox}
    \vspace{-3mm}
    \caption{\textbf{Results on PandaSet.}
    We evaluate all models on PandaSet real 64-beam data, and perform sparse-to-dense during inference optionally with ContComp and our model.
    }
    \vspace{-5mm}
    \label{tab:pandaset_pandaset}
  \end{table}

%% file: experiments-new.tex
\input{tables/crossdomain_pandaset_kitti.tex}

\vspace{-1mm}
\section{Experiments}
\label{sec:exp}

\subsection{Perception with LiDAR Completion}
\label{exp:scene_completion}
\vspace*{-2mm}
We first conduct experiments to verify that \ourmodel{} can boost the performance of a  3D detection model performance by performing LiDAR completion and using the densified results as detection model input.
To show that our model generalizes well and can generate reliable sparse-to-dense results across datasets with the learned discrete representations, we further test the models on KITTI~\cite{geiger2012we} that are trained on PandaSet~\cite{xiao2021pandaset}.
We also evaluate our model on the SemanticKITTI scene completion benchmark~\cite{behley2019semantickitti} and show that our model with a generic design can be on par with models on the leaderboard that are heavily tuned for this task.

\vspace*{-4mm}
\paragraph*{Dataset:} We train \ourmodel{} and the 3D detection models on PandaSet~\cite{xiao2021pandaset}.
We reimplement LiDARsim~\cite{manivasagam2020lidarsim} to generate the 512-beam LiDAR data paired with the real 64-beam LiDAR sweeps (\ie, the actor placement and background are identical) and obtain the sparse-to-dense supervision.
We use a custom train/validation split since there is no official split; we refer the readers to the supp. materials for more details.
We use KITTI~\cite{geiger2012we} dataset for cross-dataset generalization evaluation; the validation set is used to obtain results, and no model is trained on it.
For SemanticKITTI, we use the official train and test split.

\vspace*{-4mm}
\paragraph*{Evaluation metrics:} To evaluate 3D detection we use the KITTI  metrics and report AP$_\mathrm{BEV}$ and AP$_\mathrm{3D}$ of the car category at IoU = 0.7.
For the SemanticKITTI scene completion benchmark, we report Intersection-over-Union (IoU), which classifies a voxel as occupied or empty.

\vspace*{-4mm}
\paragraph*{Training and implementation details:} We use two base detectors for the 3D detection benchmark: PointPillar~\cite{lang2019pointpillars} from mmdetection3d~\cite{mmdet3d2020}, and a BEV detector which we denote \twostage{},  which improves over~\cite{yang2018pixor} by adding a second stage RoI refinement branch;
more details of \twostage{} can be found in the supp. materials.
We also modified PointPillar to make it use binary voxels instead of pillar features as input; so that the LiDAR completion results from our model can be used.
The voxel size for detectors is set to be the same as \ourmodel{}.
For SemanticKITTI, the input data is preprocessed as voxel data, and we directly use it without modification.

\vspace*{-5mm}
\paragraph*{Baselines: }
We first train detectors on PandaSet real data with 64-beam and simulated data with 512-beam, denoted as Real/64 and Sim/512, respectively;
during inference, we perform sparse-to-dense LiDAR completion to densify the sparse real data for the Sim/512 model.
We also train a LiDAR completion model without the codebook and quantization function (the model thus behaves like an AutoEncoder with sparse-to-dense supervision) as another baseline, denoted as ContComp.
We evaluate all models with PandaSet real data; when testing the generalization ability on KITTI dataset, we directly evaluate all models trained on PandaSet in zero-shot (\ie, no KITTI data is seen/used before the evaluation) with KITTI real data.

\vspace*{-5mm}
\paragraph{PandaSet results:}

From the first two rows in Tab.~\ref{tab:pandaset_pandaset}, we can see that the Sim/512 models perform worse than their Real/64 counterparts when evaluating on the real data.
This is expected as there is a domain gap between the dense and sparse data, and the models may learn to rely on the richer geometry information in the dense data that does not exist in the sparse counterpart.
However, if we perform the sparse-to-dense operation and lift the model to a higher density, we can see a decent performance improvement over the Sim/512 model.
Moreover, we see that our model performs better than the ContComp model that uses continuous representations, indicating the effectiveness of using discrete representation in this task.

\input{tables/semantic_kitti.tex}

\vspace*{-5mm}
\paragraph{Cross-dataset genearlization results:}
To verify the generalization ability of our model, we directly test it on KITTI in a {\it zero-shot} manner, that is without seeing KITTI data in advance.
As shown in Tab.~\ref{tab:cross_pandaset_kitti},
both entries that utilize sparse-to-dense achieve better performance, and even surpass the Real/64 model, showing the benefits of sparse-to-dense completion.
Since the discrete representation is naturally more robust to noise because of the quantization operation, our model with discrete representations can outperform ContComp by a very significant margin when doing zero-shot generalization, once again showing the benefits of using discrete representations.

\vspace*{-5mm}
\paragraph{SemanticKITTI results:}
We further test our model on SemanticKITTI scene completion benchmark. %
As shown in Tab.~\ref{tab:semantickitti}, our model can achieve on-par performance with state-of-the-art methods ~\cite{yan2021sparse,rist2021semantic,roldao2020lmscnet} that employ architectures and operations heavily tuned for this task, for example shape-aware point-voxel interaction~\cite{yan2021sparse} or multi-resolution scene representation~\cite{rist2021semantic,roldao2020lmscnet}.

\input{tables/kitti360_quantitative.tex}

\vspace{-1mm}
\subsection{LiDAR Scene Generation}
\label{exp:scene_generation}

We first compare quantitative results on KITTI-360~\cite{Liao2021ARXIV} with other baselines in the unconditional generation setting. %
In this case, we directly train our model with the sparse-to-sparse reconstruction task without the dense encoder; and use the sparse encoder to generate code maps in the generative model training instead.
We follow LiDARGen~\cite{zyrianov2022lidargen} and use the first two sequences as the validation set and use the rest for model training.
In Fig.~\ref{fig:kitti360_comparison}\footnote{We contacted the authors of~\cite{zyrianov2022lidargen} to obtain outputs of all models they used for visualization and metrics calculation}, we show the qualitative comparison between our model and two baselines; and include the real data for reference.
We can see that our model can generate results that are much more similar to the real data, with more structured and reasonable scene layouts and more stable/clearer beam patterns.
More unconditional generation results on KITTI-360 can be found in the second row of Fig.~\ref{fig:teaser} and supp. materials.

\vspace*{-3mm}
\paragraph{Quantitative results:}
Following~\cite{zyrianov2022lidargen}, we use the Maximum-Mean Discrepancy (MMD) and Jensen–Shannon divergence (JSD)  with a $100 \times 100$ 2D histogram along the ground plane (x and y coordinates) as metrics.
Since our model generates points based on voxels, the number of points may differ from the real point cloud.
We thus use occupancy as a measurement when doing histogram bin count (\ie, points from the same voxel will only count once) for point clouds from real data and other baselines.
We believe this change captures the global structure difference better and lowers the weight on the local point density estimation, which is more reasonable and aligns better with the perceptual quality.
As shown in Tab.~\ref{tab:kitti360_metrics},
our method achieves superior performance compared with the baselines.

\vspace*{-3mm}
\paragraph*{Model parameters:} We calculate the number of parameters to make sure the model capacities are at the same level when compared with baselines.
The number of parameters of LiDARGen~\cite{zyrianov2022lidargen} and our model are 29.7M and 40.3M, which are both smaller than a standard Swin-Small model.

\vspace*{-3mm}
\paragraph{Human study:} We perform an A/B test on a set of 8 researchers who have LiDAR expertise to better evaluate the visual quality of the generated samples.
We use the same test system as~\cite{zyrianov2022lidargen}, which shows a pair of randomly chosen images of two point clouds each time and lets the human decide which one is more realistic.
We compare with four baselines as well as with the real data, with  200 image pairs each, leading to 1000 image pairs in total.
We show the percentage of examples where participants believed our generations are more realistic against other baselines in Tab.~\ref{tab:human_study}.
It is clear that our model can generate results with superior visual quality; over 98.5\% of the time (100\% for some baselines), the testers prefer our results over the baselines.
It is worth noting that in 32\% cases, people believe our results are {\it more realistic than real data}, which is very significant given the fact that for data that is indistinguishable from real, the winning ratio would be 50\% with random choice.

\input{tables/human}

\vspace{-3mm}
\paragraph{Unconditional generation:}

Besides the KITTI-360 results in Fig.~\ref{fig:kitti360_comparison} and~\ref{fig:teaser}, we also train our model on dense Panadset for dense point cloud generation; and additionally run detection models on the generated samples, shown in Fig.~\ref{fig:pandaset_uncond}.
We can see that our model is able to generate diverse scenes with proper actor placement (\eg, parked car in the right example), indicating the superiority of our model for this LiDAR generation task.
Moreover, the excellent detection results showcase that the generated samples also have high fidelity w.r.t.\ the perception model.

\input{figures/pandaset_uncond.tex}
\vspace{-3mm}

\paragraph{Conditional generation:}
We show conditional generation results on KITTI-360 in the bottom row of Fig.~\ref{fig:teaser}.
Our model can fully exploit the visible part(colored by purple) as context, do reasonable extrapolation for the surrounding environment, and generate diverse scenes (\eg, curved road or crossroad) that align with the input condition well.
See supp. materials for more results on KITTI-360/PandaSet.

Meanwhile, we further consider a practical setting where the LiDAR sensor can fail for a specific region due to dirt or mechanical issues. To mimic this situation, we mask a part of the range images, as shown in Fig.~\ref{fig:rv_completion}.
We can see that our model can still do accurate completion even on partially visible cars, recovering the occluded region, and potentially avoiding dangerous situations.

\input{figures/rv_completion.tex}

\vspace{-3mm}

\paragraph{Manipulation:}
We show manipulation results with actor insertion and removal in the second row of Fig.~\ref{fig:teaser}.
This is achieved by explicitly changing the code indices on the code map and letting the decoder generate new results.
For example, we can copy the codes for the ground plane and paste them to overwrite the region where a car exists, resulting in a controllable manipulation process.
We refer the readers for more results in supp. materials.

%% file: tables/crossdomain_pandaset_kitti.tex
\begin{table}
    \centering
    \begin{adjustbox}{width=0.9\linewidth}
      \begin{tabular}{cc|cc|cc}
        \multirow{2}{*}{Model} & \multirowcell{2}{Sparse to \\ Dense}& \multicolumn{2}{c|}{\twostage{}} & \multicolumn{2}{c}{PointPillar} \\
       & &$\mathrm{AP}_{\mathrm{BEV}}$ & $\mathrm{AP}_{\mathrm{3D}}$ & $\mathrm{AP}_{\mathrm{BEV}}$  & $\mathrm{AP}_{\mathrm{3D}}$ \\
        \hline
        Real / 64 & - & 71.7 & 32.8 & 60.9 & 28.1 \\
        Sim / 512 & - & 66.9 & 33.2 & 58.5 & 28.0 \\
        Sim / 512 & ContComp& 74.9 & 41.5 & 67.7 & 36.9\\
        Sim / 512 & Ours& {\bf 76.7} & {\bf 46.3} & {\bf 73.0} & {\bf 40.9}\\
        \hline
      \end{tabular}
    \end{adjustbox}
    \vspace{-3mm}
    \caption{\textbf{PandaSet $\rightarrow$ KITTI Cross-Dataset Evaluation.}
    The detection and LiDAR completion models are trained on PandaSet, and no KITTI data is seen/used before evaluation.
    Our LiDAR completion model can do reliable completion on unseen data and significantly improve the detector performance in zero-shot.
    }
    \vspace{-5mm}
    \label{tab:cross_pandaset_kitti}
  \end{table}

%% file: tables/semantic_kitti.tex
\begin{table}[t!]
    \begin{small}
      \begin{center}
        \scalebox{0.9}{
        \begin{tabular}[c]{lcccc}
          \hline
          & SCCNet~\cite{song2017semantic} &  JS3C-Net~\cite{yan2021sparse} & Local-DIFs\cite{rist2021semantic} & Ours \\
          \hline
          IoU & 29.8 & 56.6 & 57.7 & 56.0 \\
          \hline
        \end{tabular}
        }
      \end{center}
    \end{small}
    \vspace{-5mm}
    \caption{\textbf{SemanticKITTI semantic scene completion results.}
    Our model with generic design can achieve on-par performance with state-of-the-art methods that employ architectures/modules heavily tuned for this task.
    \vspace{-6mm}
  }
    \label{tab:semantickitti}
\end{table}

%% file: tables/kitti360_quantitative.tex
\begin{table}
    \begin{small}
      \begin{center}
        \begin{tabular}[c]{c|c|c}
          \hline
          \multicolumn{1}{c|}{Method} & MMD$_{\mathrm{BEV}}\downarrow$ & JSD$_{\mathrm{BEV}}\downarrow$ \\
          \hline
          LiDAR VAE~\cite{caccia2019deep} & $1.18 \times 10^{-3}$ & 0.256\\
          LiDAR GAN~\cite{caccia2019deep} & $2.07 \times 10^{-3}$ & 0.275\\
          Projected GAN~\cite{sauer2021projected} & $1.25 \times 10^{-3}$ & 0.190\\
          LiDARGen~\cite{zyrianov2022lidargen} & $4.80 \times 10^{-4}$ & 0.140 \\
          Ours & ${\bf 9.67 \times 10^{-5}}$ & {\bf 0.132} \\
          \hline
        \end{tabular}
      \end{center}
    \end{small}
    \vspace{-5mm}
    \caption{{\bf Quantitative results on KITTI-360.} Our results show better statistical alignment with the real data.
    \vspace{-5mm}
    }
    \label{tab:kitti360_metrics}
  \end{table}

%% file: tables/human.tex
  \begin{table}[t]
    \begin{small}
      \begin{center}
        \begin{tabular}[c]{c|c}
          \hline
          \multicolumn{1}{c|}{Method} & Percent Prefer Ours \\
          \hline
          Ours vs LiDAR VAE~\cite{caccia2019deep} & $99.5\%$ \\
          Ours vs LiDAR GAN~\cite{caccia2019deep} & $100\%$\\
          Ours vs ProjGAN~\cite{sauer2021projected} & $100\%$ \\
          Ours vs LiDARGen~\cite{zyrianov2022lidargen} & $98.5\%$\\
          \hline
        \end{tabular}
      \end{center}
    \end{small}
    \vspace{-6mm}
    \caption{{\bf Human study results on KITTI-360.}
    Results from our model show significantly better visual quality.
    }
    \label{tab:human_study}
  \end{table}

%% file: figures/pandaset_uncond.tex
\begin{figure}
    \centering
    \vspace{-3mm}
    \includegraphics[width=\linewidth]{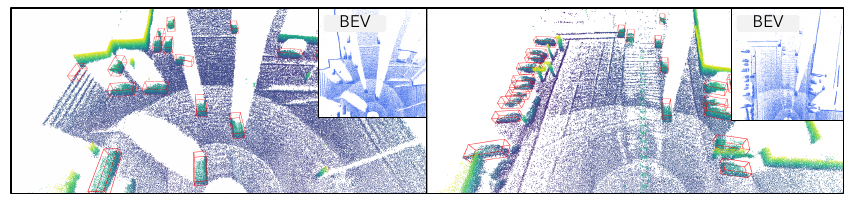}
    \vspace{-6mm}
    \caption{\textbf{Unconditional generation results on Pandaset.} We train our model on dense Pandaset data and generate dense results.
    The generated samples show diverse scenario layouts with proper actor placement (\eg, parked car in the right sample).
    The synthesized point clouds are realistic such that a pre-trained detector can directly work out-of-the-box.
    }\label{fig:pandaset_uncond}
    \vspace{-6mm}
\end{figure}

%% file: figures/rv_completion.tex
\begin{figure}
    \includegraphics[width=\linewidth]{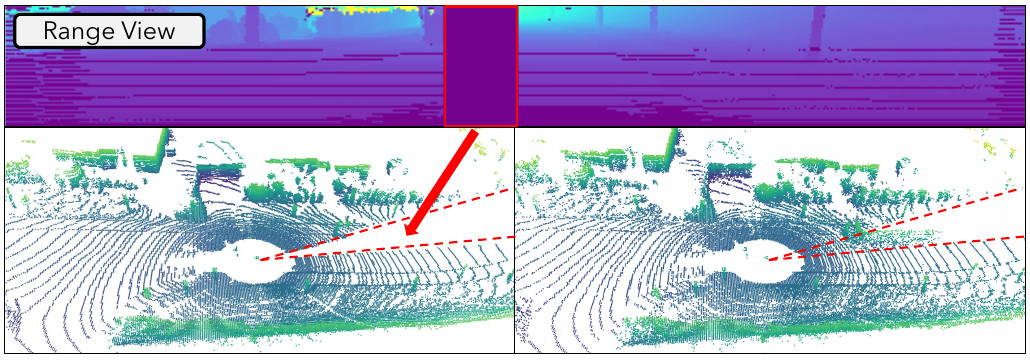}
    \vspace{-6mm}
    \caption{\textbf{Conditional LiDAR generation for dirt removal.}
    We mask the red rectangular region in the range view image to mimic dirt occluder.
    \textbf{Left}: Original input. The masked region is not visible to the model.
    \textbf{Right:} Our generation results.
    Our model can successfully recover the vehicle that is partially observed.}\label{fig:rv_completion}
	\vspace{-5mm}
\end{figure}

%% file: conclusion.tex
\vspace{-2mm}
\section{Conclusion}
\vspace{-2mm}
In this paper, we propose a framework to learn novel LiDAR representations that effectively capture the data priors that enable various downstream applications, including scene completion and generation.
Based on the learned representation, we propose a novel sparse-to-dense data reconstruction pipeline that can accurately densify the sparse input and improve the generalization ability of the trained detection models in a zero-shot manner.
We further show that our model can generate high-quality LiDAR point clouds unconditionally and do reasonable conditional generation and manipulation, which leads to more realistic and controllable data-driven LiDAR simulation.

One potential future direction for improving our model is to exploit the structure in range view representations, including beam info (which point from which ray) and occlusion reasoning, and combine them with our existing model.
This may potentially improve the fidelity of our results.

%% file: acknowledgement.tex
\vspace{-4mm}
\paragraph*{Acknowledgement}

We thank Vlas Zyrianov and Shenlong Wang for providing A/B test code and baseline results, Arnaud Bonnet for infrastructure support, Siva Manivasagam for proofreading.
We also thank the Waabi team for their valuable support.
YX is partially supported by an NSERC scholarship and a Borealis AI fellowship.
WCM is partially supported by a Siebel scholarship.

%% file: supp.tex
\section{Code Identification}

For street scenes in Birds-Eye-View (BEV), we empirically observe that the semantic meaning of the codes aligned well with the actors spatially.
For example, we can refer object (\eg, car) location from ground-truth labels or visual check and find the corresponding codes on the code map.
The meaning of the codes can then be identified for controllable generation and manipulation.
In this paper, we mainly leverage this property to perform scene manipulation by using a copy-paste mechanism for existing code and find it works properly.
How to find the meaning of the codes for background/unlabeled effectively is still an open question.
We leave this question and more potential use cases for the identified codes as future works.

\section{Experimental Details}

In this section, we discuss the experimental details for the results presented in the main paper.

\paragraph{PanadSet:}
PandaSet is a self-driving dataset introduced in recent work~\cite{xiao2021pandaset}.
It contains 103 driving sequences captured in San Francisco Bay area with 8 seconds (80 frames, sampled at 10Hz) each.
We split it into 73 and 30 sequences as training and validation sets by considering both geographic locations and time (so that data collected at different locations/timespan are in different sets).
Specifically, we select the sequences $13-14, 57-79, 86-94$ and $149$ as the validation set and put the rest into the training set.
We set the region of interest for the point cloud to $[0, 80] \times [-40, 40]$ meters.

\paragraph{\twostage{}:}
\twostage{} can be seen as a variant of the original PIXOR proposed by Yang \etal~\cite{yang2018pixor}.
We made the following modifications to improve the performance to be comparable with the state-of-the-art 3D detectors:

\begin{enumerate}
    \item We use multi-scale deformable self-attention~\cite{zhu2020deformable} instead of the upsampling deconvolution layers after the ResNet backbone to aggregate information from different scales and enhance the feature extraction; and output the feature maps with $1/4$ resolution the same as PIXOR. %
    \item We use the output from the dense detector header as the first-stage results. The top 500 bounding boxes after NMS are used as region proposals for the second stage. The 2D IoU threshold for NMS is set to 0.7
    \item We use RotatedRoIAlign~\cite{wu2019detectron2} to extract $3\times 3$ RoI features from the feature map in the second-stage header; and apply two self-attention layers on features within each RoI and features between each RoI, respectively.
Two MLPs are used to predict the classification score and box refinement for the region proposals.
    \item We use DETR-like set-based loss with bipartite matching for both stages. An additional IoU loss is used for bounding box regression besides the smooth L1 loss.
\end{enumerate}

\paragraph{LiDAR Simulation:}
We re-implement the state-of-the-art LiDAR simulation approach LiDARsim~\cite{manivasagam2020lidarsim} for the generation of sim data with 512 laser beams. We only focus on the physics based simulation component based on OptiX ray tracing engine~\cite{parker2010optix} and skip the ML ray-drop network.

We first create the asset bank with log-wise surfel aggregation following~\cite{manivasagam2020lidarsim}. For background assets, we aggregate the points across all the frames in one snippet and remove all actors using  3D bounding box annotations. For dynamic actors, we aggregate the LiDAR points inside the bounding boxes in the object-centric coordinate. We then estimate per-point normals from 200 nearest neighbors with a radius of $20cm$ and orient the normals upwards for flat ground reconstruction. We downsample the LiDAR points into 4cm voxels and create per-point triangle faces (radius $5cm$)  according to the estimated normals. Given the asset bank, we place the background and all actors in its original locations and transform the scene to the LiDAR coordinate for ray-triangle intersection computation.  For Pandaset, we set the sensor intrinsics (\eg, beam angle, azimuth resolution, etc) based on the public documents and raw LiDAR information~\footnote{\url{https://github.com/scaleapi/pandaset-devkit/issues/67}}. To produce simulated LiDAR scans with 512 beams, we linearly interpolate the beam angle from 14.870\textdegree \ to -30.166\textdegree, where laser ID $0, 8, \cdots 504$ corresponds to the original 64 beams (14.870\textdegree\ to -24.909\textdegree) for Pandar64. To reduce the domain gap between simulated 512-beam LiDAR data and real data, we replace simulated 64-beam  LiDAR points (laser IDs $0, 8, \cdots 504$, 64 beams in total) with the original real LiDAR points.

\paragraph{Metrics on KITTI-360:}

The Maximum-Mean Discrepancy (MMD) and Jensen–Shannon divergence (JSD) metrics~\cite{zyrianov2022lidargen} measure the distribution between the generated results and the real data with a $100\times 100$ 2D histogram along the ground plane.
In the main paper, we use occupancy as the measurement for all generation methods when doing histogram bin count for point clouds from real data.
We now try to measure the point-based distribution by lifting our results with point duplication.
Specifically, we calculate a coefficient for each bin used for the histogram bin count by performing an element-wise division of the 2D histogram matrix calculated on KITTI-360 training data with the histogram calculated on our generated data.
We use the coefficients to ``duplicate'' the points in each voxel and match the point distribution.
Our method still performs well in this case, as shown in Tab.~\ref{tab:kitti360_metrics_supp}.

\input{tables/supp/kitti360_quantitative_supp.tex}
\input{figures/supp/kitti360_no_freespace_supp.tex}
\input{figures/supp/kitti360_uncond_16x.tex}
\input{figures/supp/kitti360_uncond_no_reinit.tex}

\section{Ablation Studies}

We now show ablation studies on the design choices described for our model.
Specifically, as it is easier to observe the difference between the generation results and the real data,
we show qualitative results on KITTI-360.

\paragraph{Free space suppression:}
In Fig.~\ref{fig:kitti360_no_freespace_supp}, we show generation results without code restriction (\ie, every code is available to choose, including the \texttt{[BLANK]} codes).
In this situation, we do not perform any free space suppression, and we can see that the generation is clearly collapsed, indicating the effectiveness of our free space suppression.

\paragraph{Voxel size:}
We empirically find that the spatial dimension that each code represents should not be too large.
In Fig.~\ref{fig:kitti360_uncond_16x}, we show qualitative results of downsampling $16\times$ for the model before vector quantization, leading to a $2.5 \times 2.5$ m patch for each code.
We can see that the results become noisy as $2.5\times 2.5$ m patches contain too many geometry details that cannot be preserved in a single discrete code.
We thus stay with $8\times$ downsampling for our model.

\paragraph{Codebook (re)-initialization:}
We also notice that the codebook initialization and reinitialization matter when learning the codebook.
We train a model with conventional uniform initialization and no reinitialization.
The code utilization rate at the end of codebook learning is 12.1\%, significantly lower than our implementation with reinitialization which can achieve 99\%+.
As the qualitative results are shown in Fig.~\ref{fig:kitti360_uncond_no_reinit}, it is clear that the limited number of active codes leads to degenerated results.

\input{figures/supp/pandaset_sparse_to_dense_supp.tex}

\section{Visualization}

We now show more visual examples on PandaSet and KITTI360.

\vspace*{-2mm}

\paragraph{Sparse-to-dense results}
We show more sparse-to-dense results on PandaSet in Fig.~\ref{fig:pandaset_sparse2dense_supp}; we also include a video for results with a full sequence in the supplementary.

\vspace*{-2mm}

\paragraph{Unconditional generation results:}
We show more results compared with baselines on KITTI-360 in Fig.~\ref{fig:kitti360_comparison_supp}.
Furthermore, we also show KITTI-360 results with the generation process in Fig.~\ref{fig:kitti360_uncond_supp}, and PandaSet results in Fig.~\ref{fig:pandaset_uncond_supp}, respectively.

\input{figures/supp/kitti360_comparison.tex}
\input{figures/supp/kitti360_uncond.tex}
\input{figures/supp/pandaset_uncond_supp.tex}

\vspace*{-2mm}

\paragraph{Conditional generation results:}
We show KITTI-360 results in Fig.~\ref{fig:kitti360_cond_supp}, and PandaSet results in Fig.~\ref{fig:pandaset_cond_supp} on PandaSet, respectively.
The red points indicate the visible part as the condition for the model, and our model can perform reasonable extrapolation based on that.
\input{figures/supp/kitti360_cond_supp.tex}
\input{figures/supp/pandaset_cond_supp.tex}

\vspace*{-2mm}

\paragraph{Manipulation results:}
Lastly, we show more manipulation results on KITTI-360 and PandaSet in Fig.~\ref{fig:manip_combine}.
The learned codes show high spatial alignment with the objects, so we can manipulate the LiDAR sweeps by changing the code placement on the code map.

\begin{figure*}
    \vspace*{-0.25cm}
    \includegraphics[width=\linewidth]{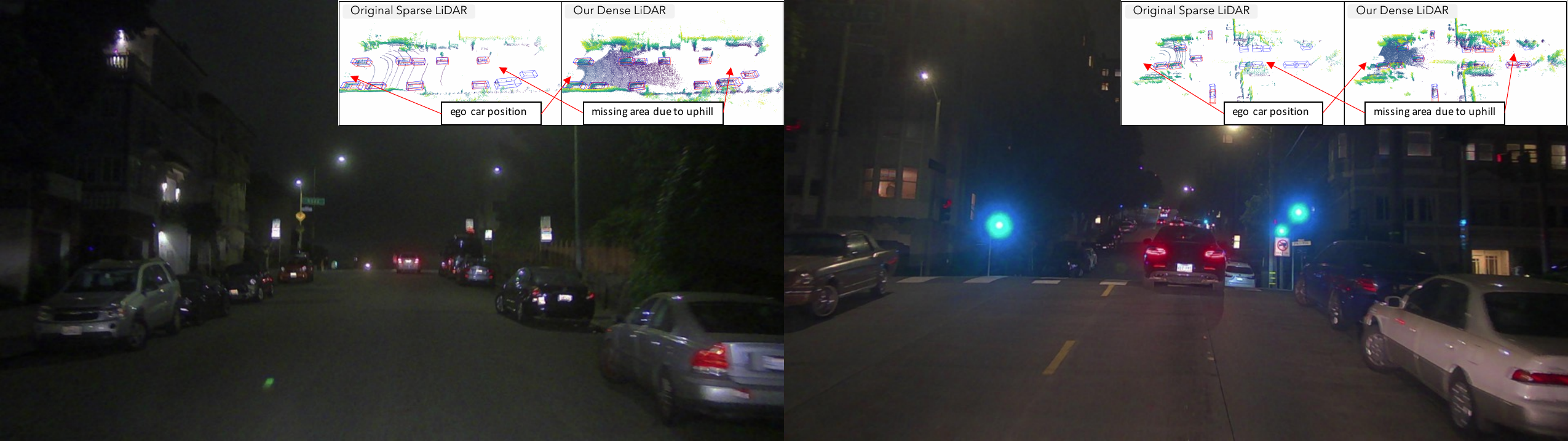}
    \caption{\small Camera images for the corresponding LiDAR completion examples. See text for detailed explanations.}
    \label{fig:sparse2dense}
\end{figure*}

\paragraph*{Missing part in completion:} UltraLiDAR is designed to densify the input point clouds while maintaining realistic occlusion patterns such that the data appear as if it were captured by a LiDAR sensor with a higher beam count.
The large missing volume in Fig. 3 in the main paper is due to uphill/downhill geometry of the road, which results in occlusion for any LiDAR regardless of their beam count. The RGB camera images shown in Fig.~\ref{fig:sparse2dense} provides a clearer illustration.  %
\vspace*{-0.45cm}

\input{figures/supp/manip_combine.tex}

%% file: tables/supp/kitti360_quantitative_supp.tex
\begin{table}
    \begin{small}
      \begin{center}
        \begin{tabular}[c]{c|c|c}
          \hline
          \multicolumn{1}{c|}{Method} & MMD$_{\mathrm{BEV}}\downarrow$ & JSD$_{\mathrm{BEV}}\downarrow$ \\
          \hline
          LiDAR GAN~\cite{caccia2019deep} & $3.06 \times 10^{-3}$ & - \\
          LiDAR VAE~\cite{caccia2019deep} & $1.00 \times 10^{-3}$ & 0.161 \\
          Projected GAN~\cite{sauer2021projected} & $3.47 \times 10^{-4}$ & 0.085\\
          LiDARGen~\cite{zyrianov2022lidargen} & $3.87 \times 10^{-4}$ & {\bf 0.067} \\
          Ours & ${\bf 1.96 \times 10^{-4}}$ & 0.071 \\
          \hline
        \end{tabular}
      \end{center}
    \end{small}
    \vspace{-5mm}
    \caption{{\bf Quantitative results on KITTI-360.} Baseline results are from~\cite{zyrianov2022lidargen}.}
    \label{tab:kitti360_metrics_supp}
  \end{table}

%% file: figures/supp/kitti360_no_freespace_supp.tex
\begin{figure*}[htbp!]
    \includegraphics[width=\linewidth]{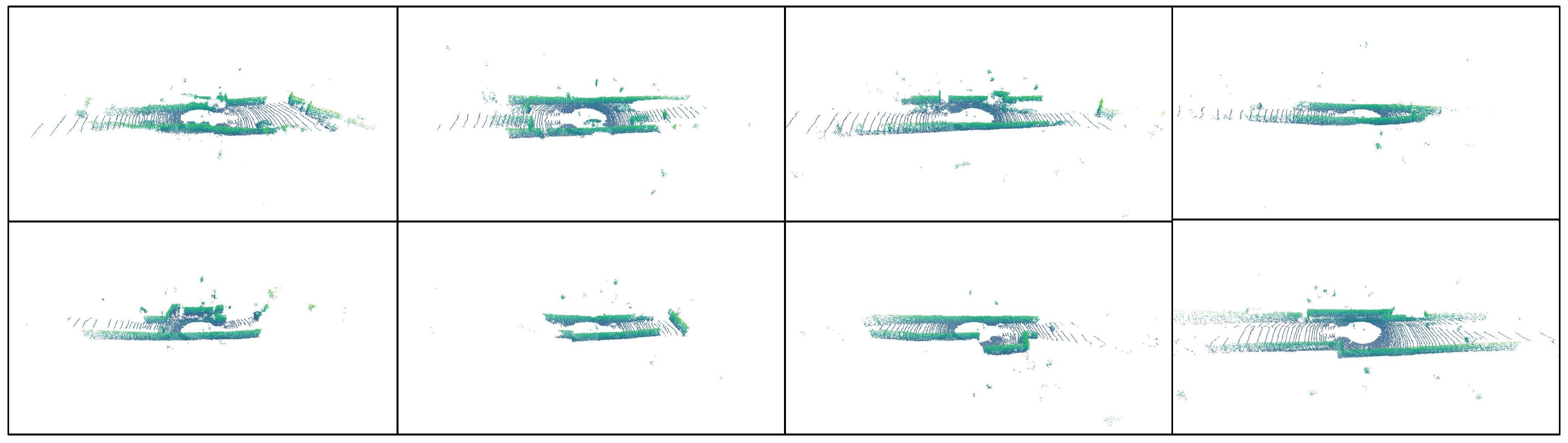}
    \caption{\textbf{Results without free space suppression}. Most space is now empty; the model can no longer generate structured layouts.}
    \label{fig:kitti360_no_freespace_supp}
  \end{figure*}

%% file: figures/supp/kitti360_uncond_16x.tex
\begin{figure*}[htbp!]
    \includegraphics[width=\linewidth]{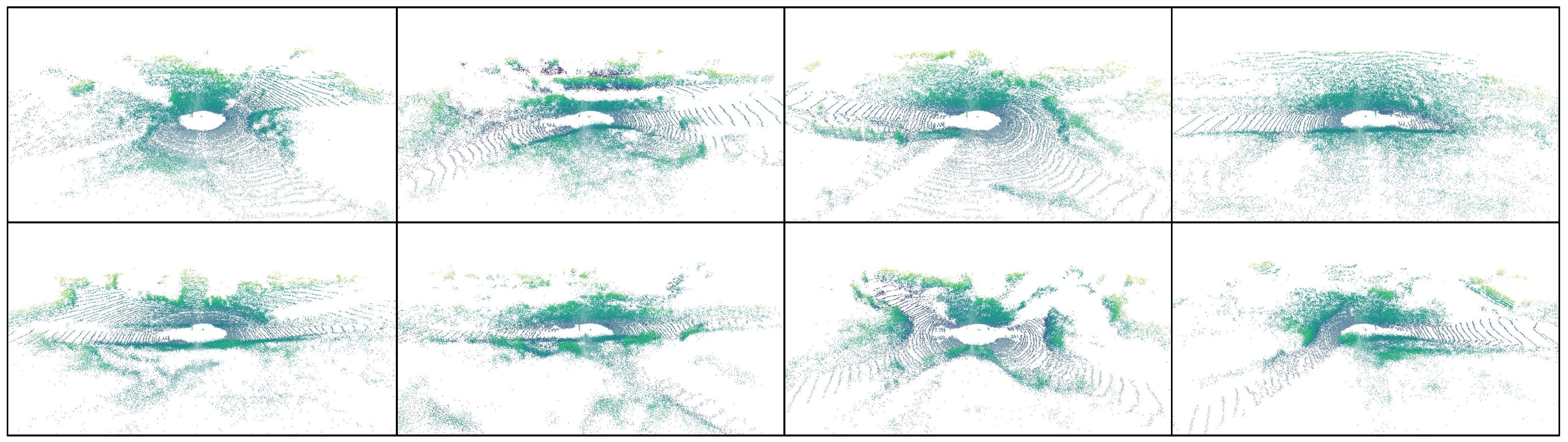}
    \caption{\textbf{Results with larger patch size for each code.} The point clouds become noisy as the patch size for each code is too large, and fine-grained geometry cannot be preserved anymore.}
    \label{fig:kitti360_uncond_16x}
  \end{figure*}

%% file: figures/supp/kitti360_uncond_no_reinit.tex
\begin{figure*}
    \includegraphics[width=\linewidth]{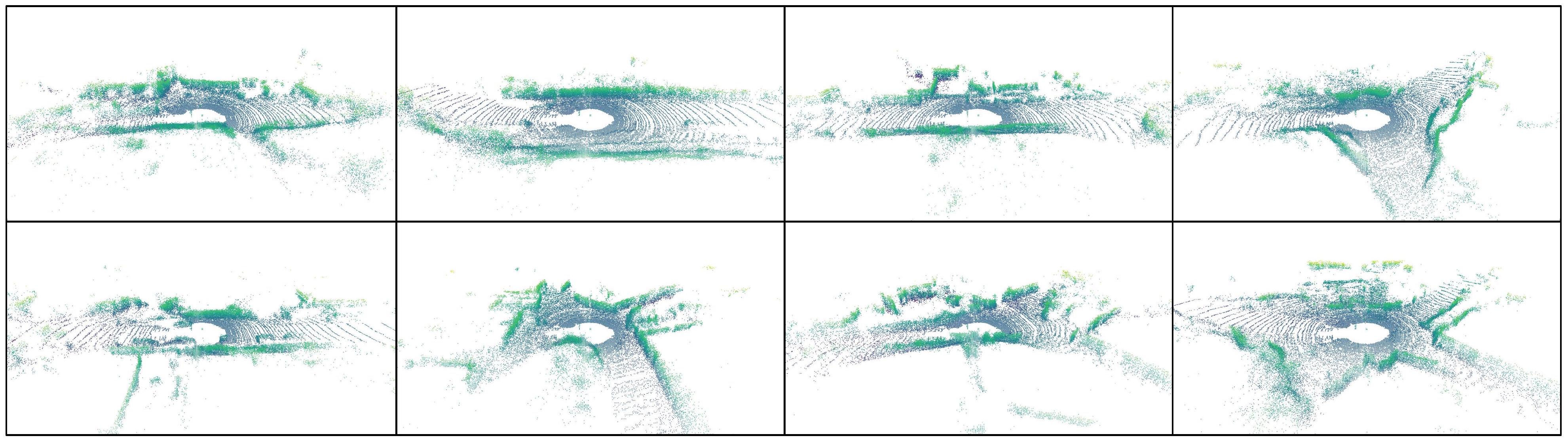}
    \caption{\textbf{Results without codebook reinitialization.} We observe similar issue as in Fig.~\ref{fig:kitti360_uncond_16x}. The model needs to use a limited number of codes to perform reconstruction/generation, leading to degenerated results.}
    \label{fig:kitti360_uncond_no_reinit}
  \end{figure*}

%% file: figures/supp/pandaset_sparse_to_dense_supp.tex
\begin{figure*}
    \includegraphics[width=\textwidth]{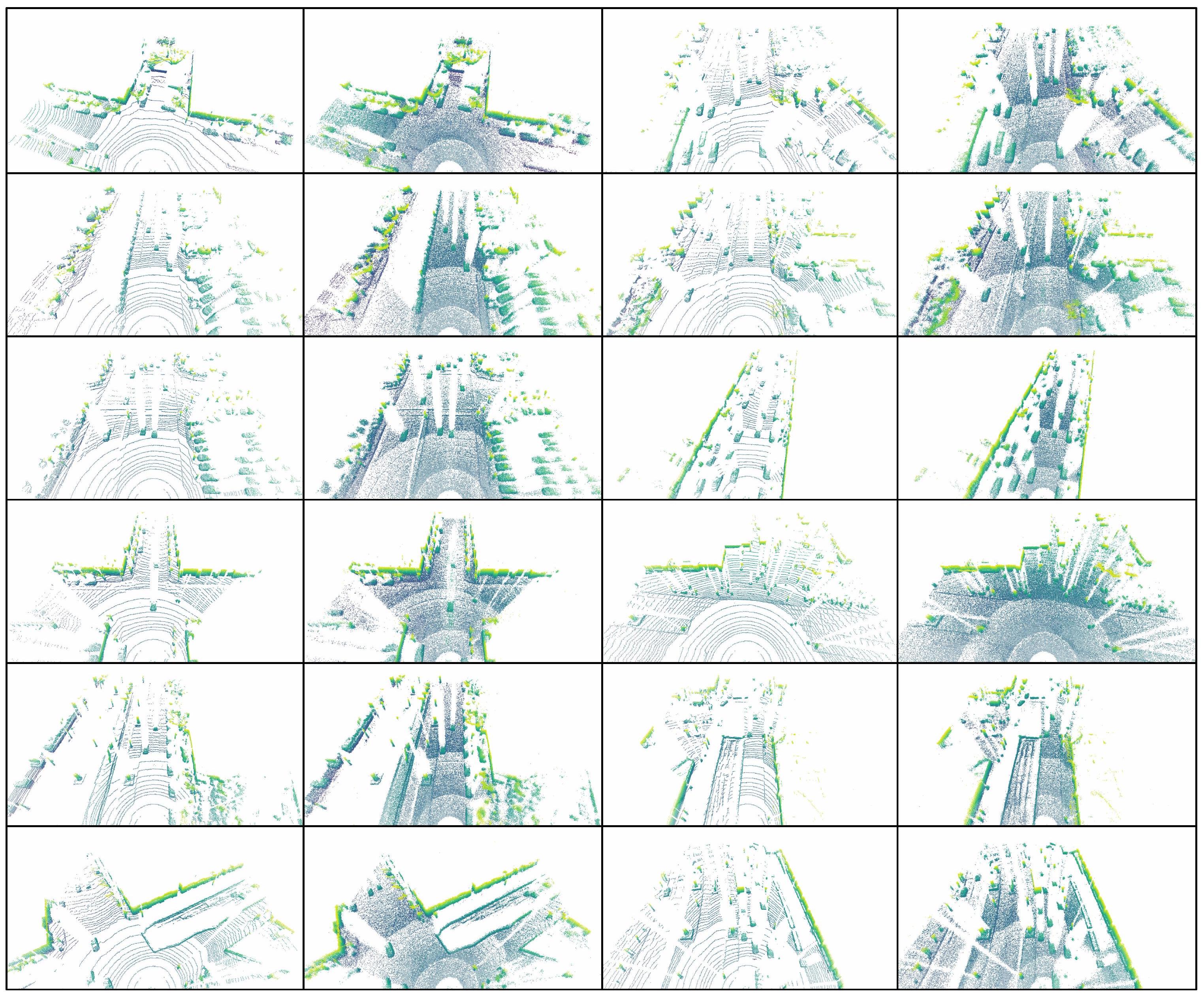}
    \caption{\textbf{Sparse-to-Dense results on PandaSet.} The first and third columns are real sparse data, and the second and fourth columns are our densified results.}
    \label{fig:pandaset_sparse2dense_supp}
  \end{figure*}

%% file: figures/supp/kitti360_comparison.tex
\begin{figure*}[t]
    \centering
    \vspace{-7mm}
    \includegraphics[width=1.0\textwidth]{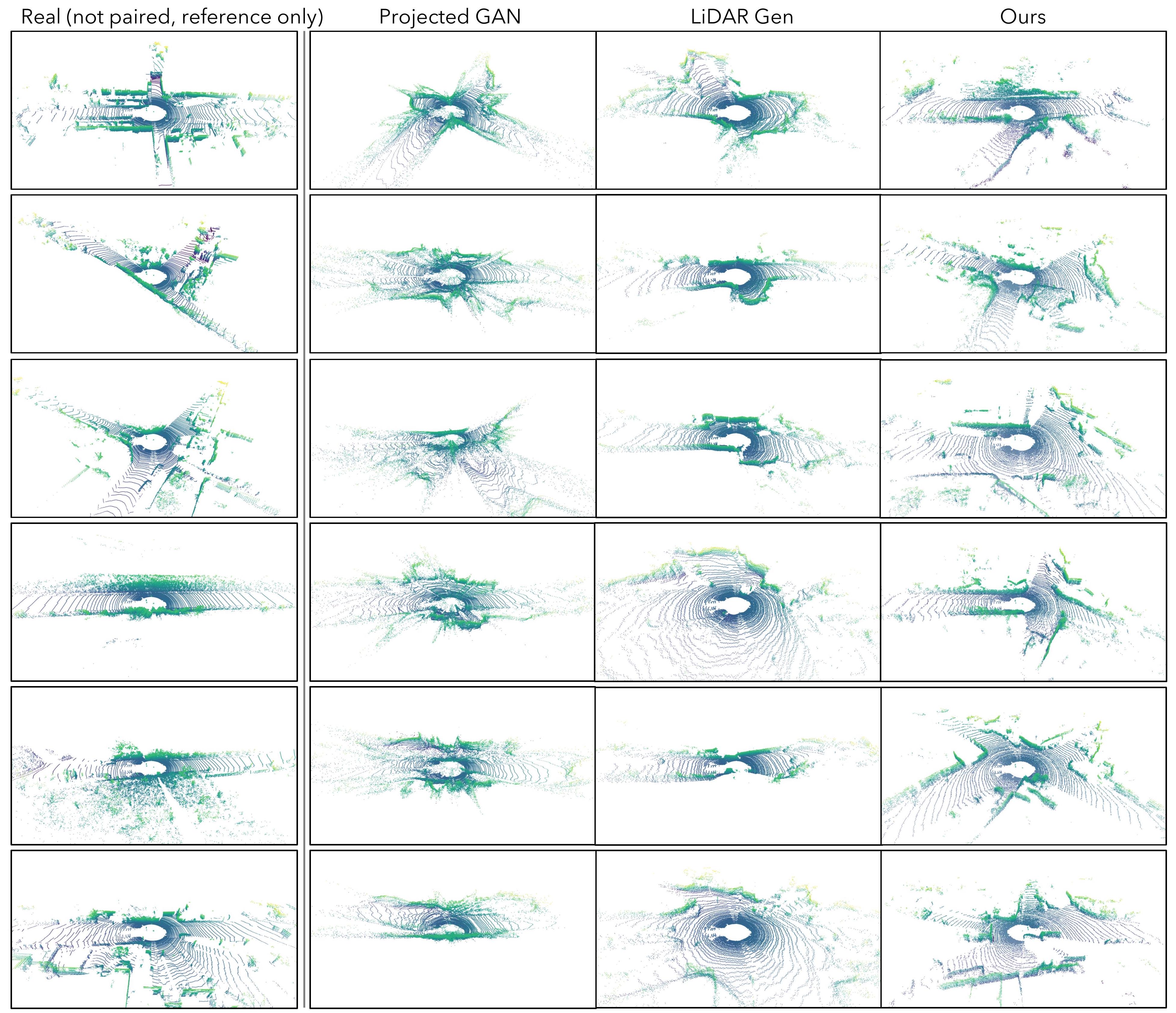}
    \vspace{-3mm}
    \caption{\textbf{Qualitative comparison against baselines on unconditional LiDAR generation.}
    Our model consistently outperforms the baselines and shows results highly similar to the real data.
    }
    \vspace{-5.5mm}
    \label{fig:kitti360_comparison_supp}
\end{figure*}

%% file: figures/supp/kitti360_uncond.tex
\begin{figure*}
    \includegraphics[width=\textwidth]{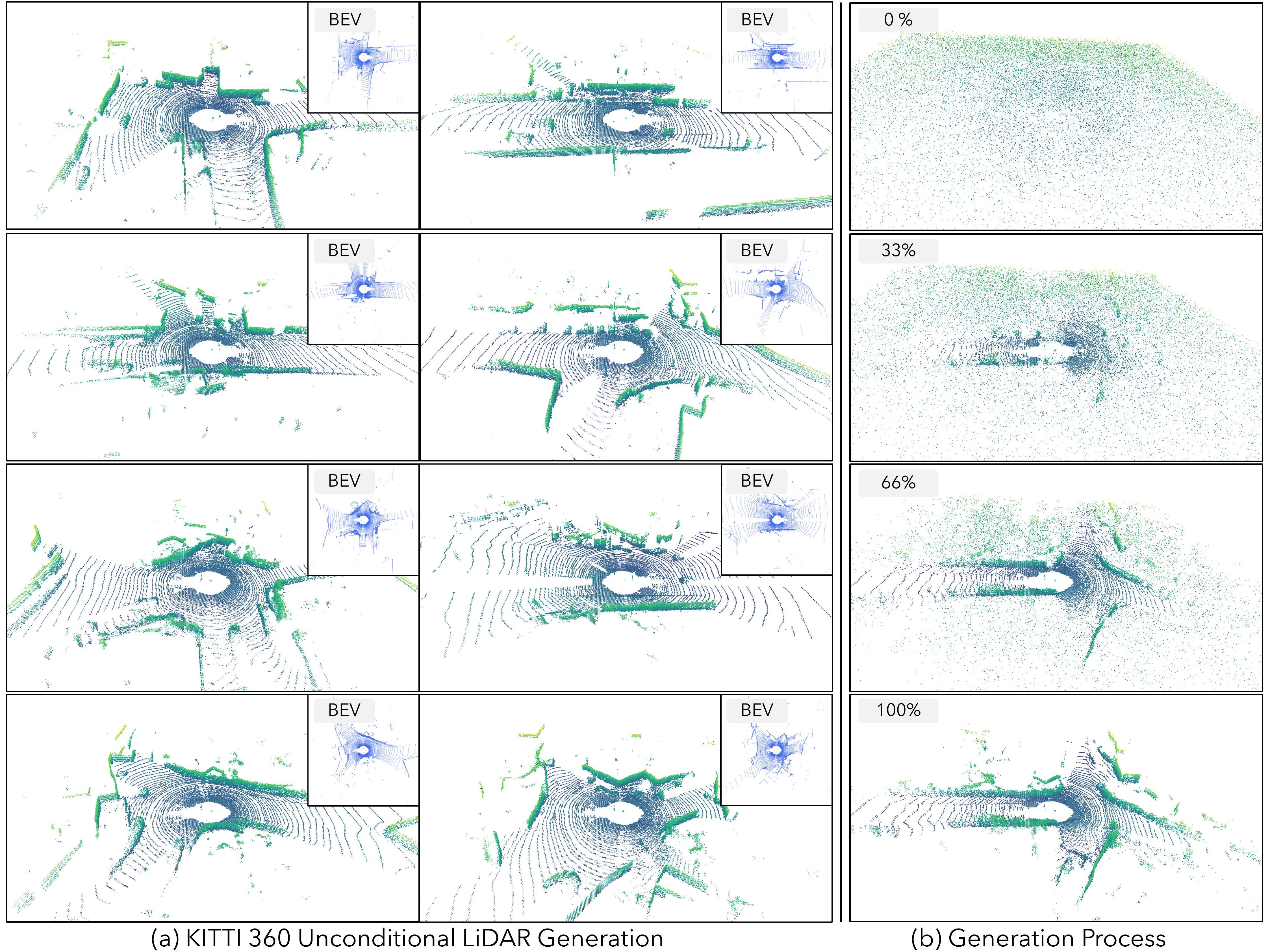}
    \caption{\textbf{Unconditional generation results on KITTI-360.} We show a step-by-step generation process starting from a blank canvas in the third column.}
    \label{fig:kitti360_uncond_supp}
  \end{figure*}

%% file: figures/supp/pandaset_uncond_supp.tex
\begin{figure*}
    \includegraphics[width=\textwidth]{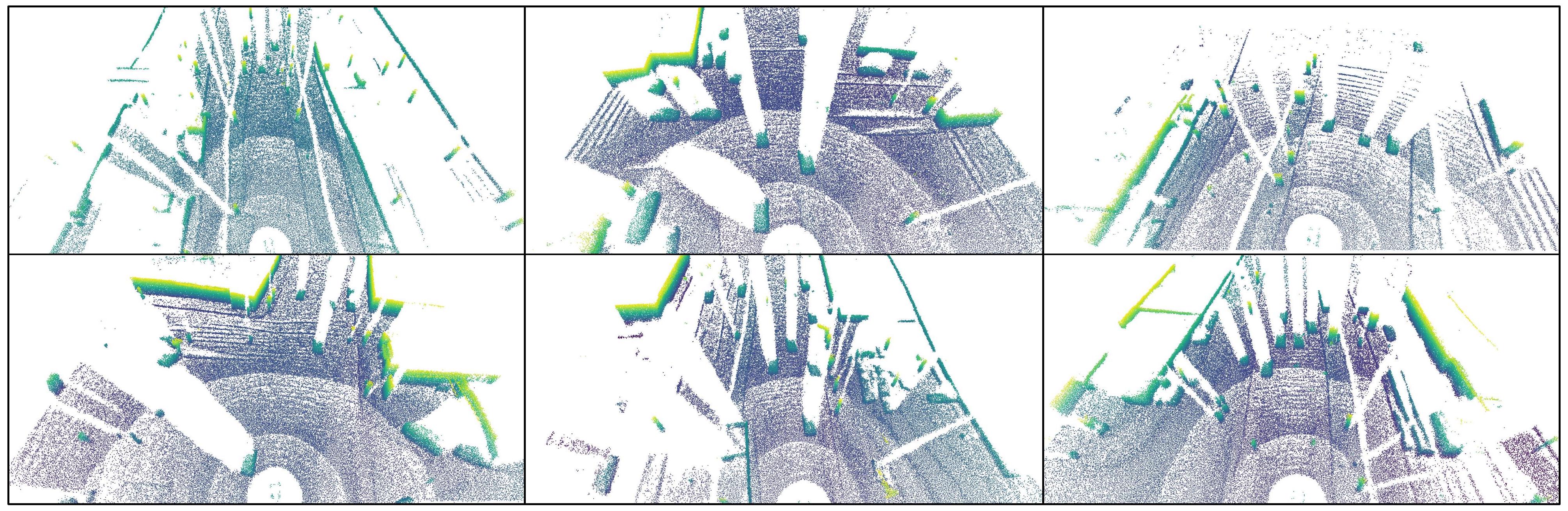}
    \caption{\textbf{Unconditional generation results on Pandaset.}}
    \label{fig:pandaset_uncond_supp}
  \end{figure*}

%% file: figures/supp/kitti360_cond_supp.tex
\begin{figure*}
    \includegraphics[width=\textwidth]{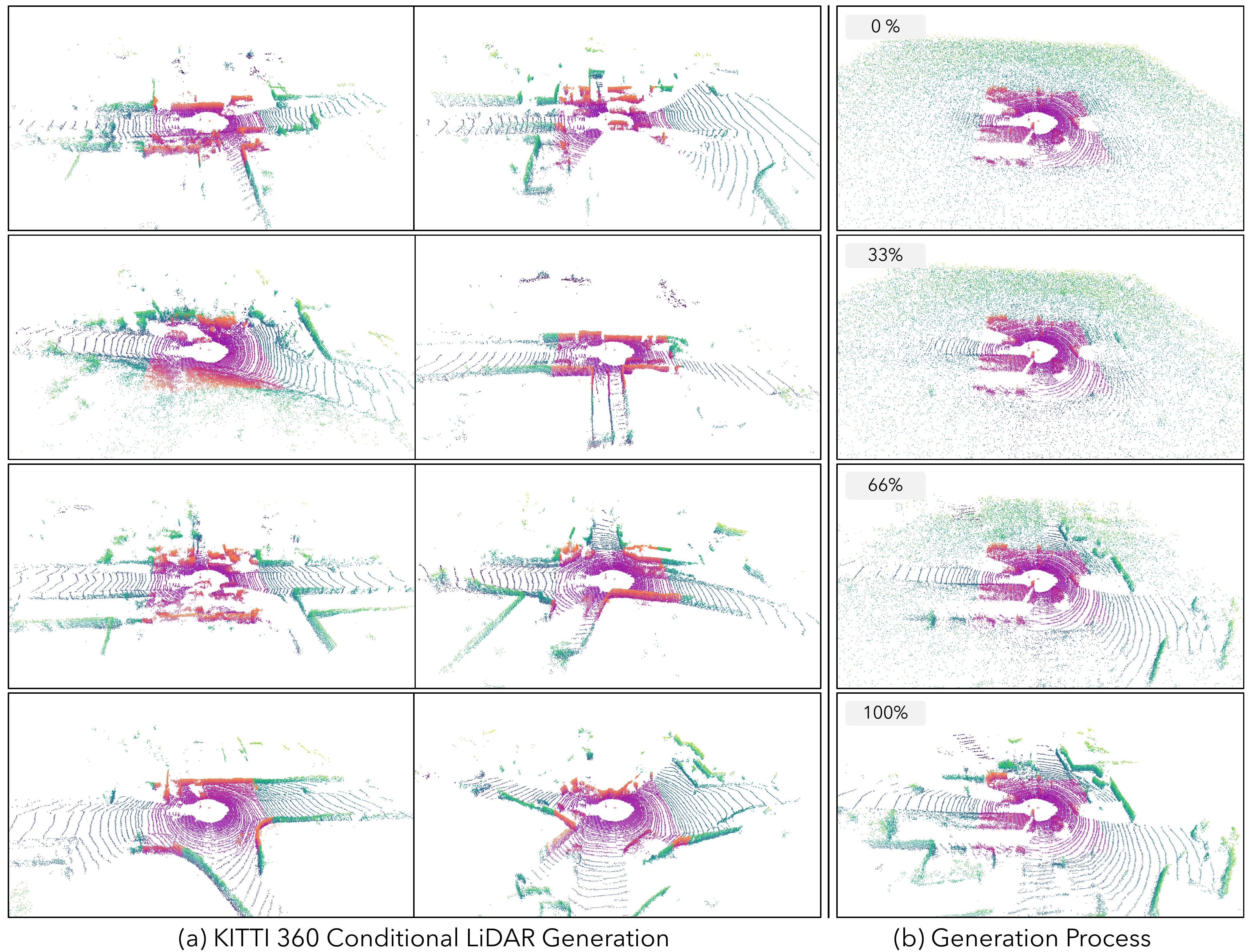}
    \caption{\textbf{Conditional generation results on KITTI-360.} The red points are the visible input to the model.}
    \label{fig:kitti360_cond_supp}
  \end{figure*}

%% file: figures/supp/pandaset_cond_supp.tex
\begin{figure*}
    \includegraphics[width=\textwidth]{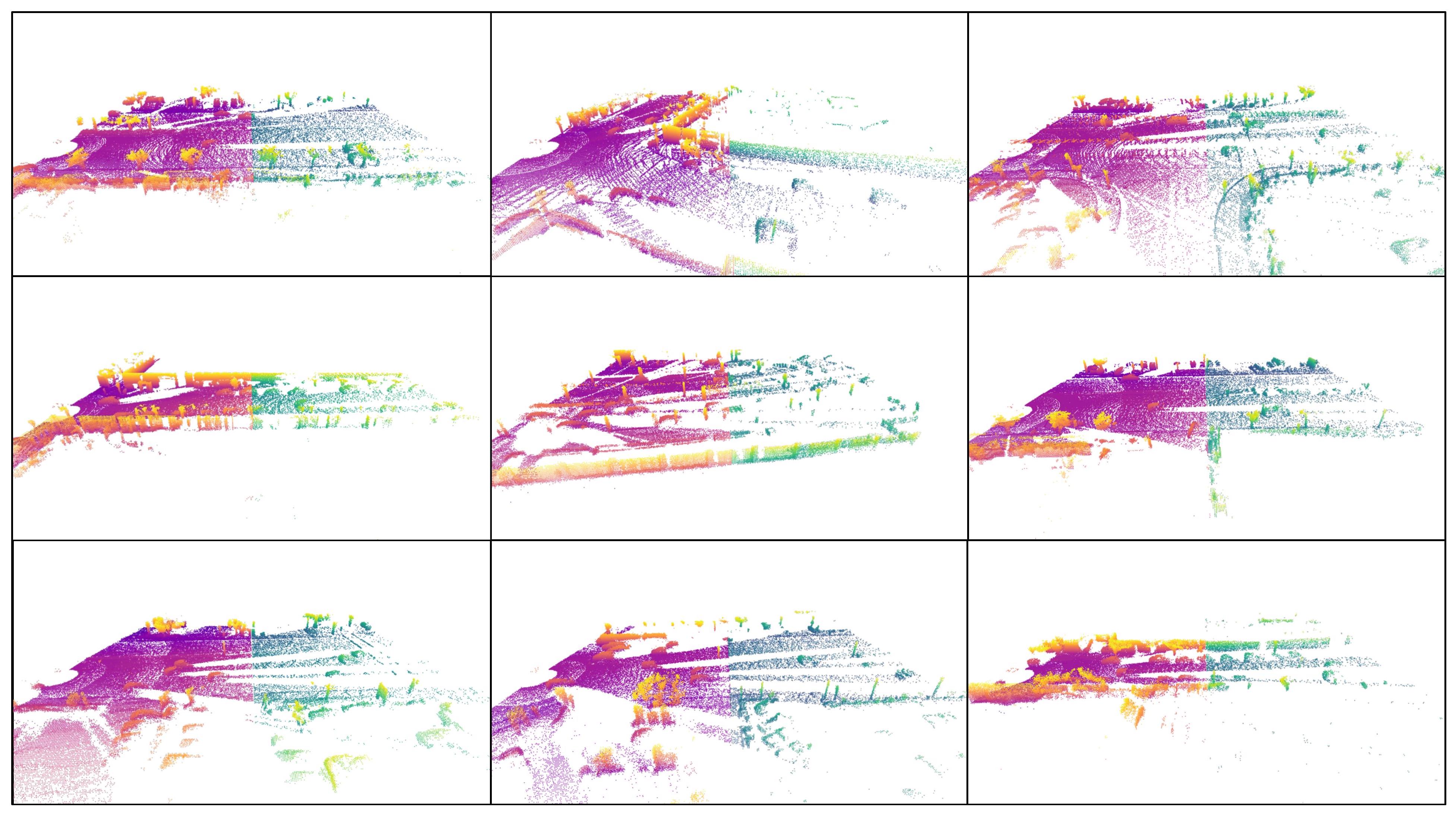}
    \caption{\textbf{Conditional generation results on Pandaset.} The red points are the visible input to the model.}
    \label{fig:pandaset_cond_supp}
  \end{figure*}

%% file: figures/supp/manip_combine.tex
\begin{figure*}
    \includegraphics[width=\linewidth]{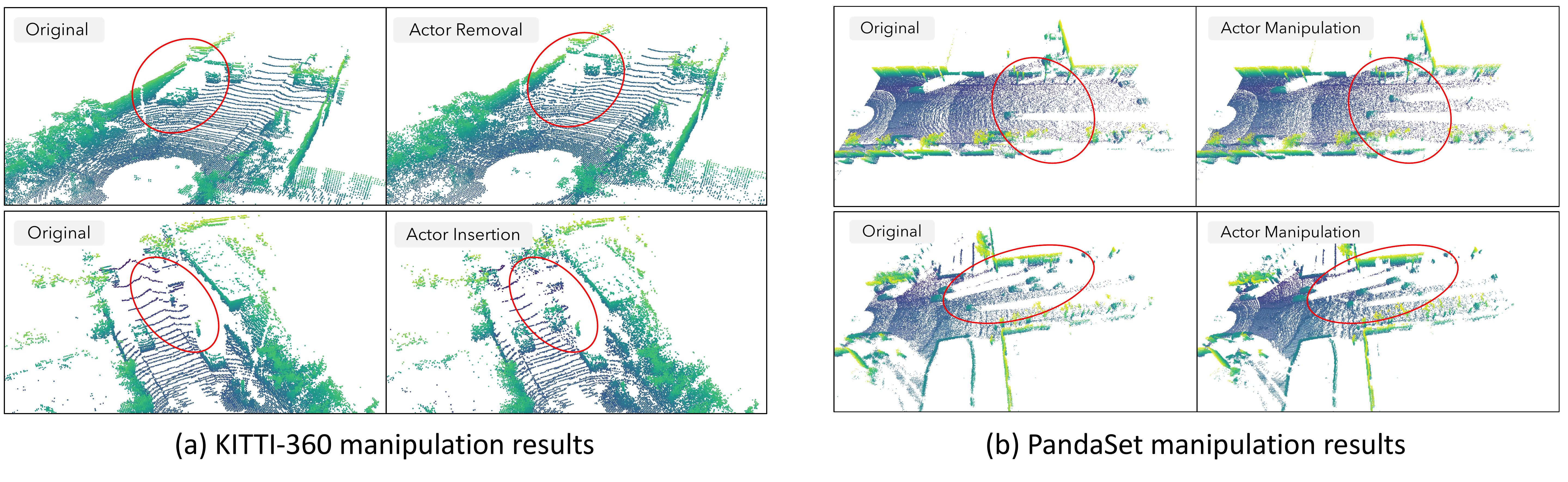}
    \caption{\textbf{Manipulation results on (a) KITTI-360 and (b) PandaSet.} The learned codes show high spatial alignment with the objects, so we can manipulate the LiDAR sweeps by changing the code placement on the code map.}
    \label{fig:manip_combine}
  \end{figure*}